\documentclass[10pt,twocolumn,letterpaper]{article}

\usepackage[pagenumbers]{cvpr} 

\usepackage{graphicx}
\usepackage{amsmath}
\usepackage{amssymb}
\usepackage{booktabs}
\usepackage{epsfig}
\usepackage{amsmath}
\usepackage{amssymb}
\usepackage{booktabs}
\usepackage{tabulary}
\usepackage{multirow}
\usepackage{overpic}
\usepackage{xspace}
\usepackage{bbm}
\usepackage{xcolor}         
\usepackage{microtype}
\usepackage{amsmath,amsfonts,bbm}
\usepackage{multirow}
\usepackage{epsfig}
\usepackage{times,amsmath,amssymb,booktabs,tabulary,multirow,overpic,xcolor}
\usepackage{mathtools}
\usepackage{wrapfig}
\usepackage{nicefrac}       
\usepackage{microtype}      
\usepackage{enumitem}
\usepackage{bbm}

\usepackage[pagebackref=false, breaklinks=true, letterpaper=true, colorlinks,
            citecolor=citecolor, linkcolor=linkcolor, bookmarks=false]{hyperref}
\definecolor{citecolor}{HTML}{0071BC}
\definecolor{linkcolor}{HTML}{ED1C24}

\definecolor{demphcolor}{RGB}{144, 144, 144}
\newcommand{\demph}[1]{\textcolor{demphcolor}{#1}}

\renewcommand{\paragraph}[1]{\vspace{1.25mm}\noindent\textbf{#1}}

\usepackage[ruled]{algorithm2e}
\usepackage{algorithmic}

\graphicspath{ {./figs/} }
\usepackage{floatrow}
\usepackage{adjustbox}
\newfloatcommand{capbtabbox}{table}[][\FBwidth]

\usepackage[capitalize]{cleveref}
\crefname{section}{Sec.}{Secs.}
\Crefname{section}{Section}{Sections}
\Crefname{table}{Table}{Tables}
\crefname{table}{Tab.}{Tabs.}

\def\paperID{} 
\def\confName{ICCV}
\def\confYear{2025}

\usepackage{multirow}
\usepackage{cuted}

\newcommand{\vct}[1]{\boldsymbol{#1}} 
\newcommand{\mat}[1]{\boldsymbol{#1}} 

\newcommand{\methodname}{{VICP}\xspace}

\newcommand{\app}{\raise.17ex\hbox{$\scriptstyle\sim$}}

\newlength\savewidth

\usepackage{overpic}

\usepackage[misc]{ifsym}

\usepackage[hang]{footmisc}
\title{
Generalizable Object Re-Identification via Visual In-Context Prompting
}

\author{
Zhizhong Huang \quad Xiaoming Liu\\
Michigan State University\\
East Lansing, MI, USA\\
{\tt\small \{huang296, liuxm\}@msu.edu}
}

\begin{document}
\maketitle
\thispagestyle{empty}

\begin{abstract}

Current object re-identification (ReID) methods train domain-specific models (e.g., for persons or vehicles), which lack generalization and demand costly labeled data for new categories. While self-supervised learning reduces annotation needs by learning instance-wise invariance, it struggles to capture \textit{identity-sensitive} features critical for ReID.
This paper proposes Visual In-Context Prompting~(\methodname), a novel framework where models trained on seen categories can directly generalize to unseen novel categories using only \textit{in-context examples} as prompts, without requiring parameter adaptation. \methodname synergizes LLMs and vision foundation models~(VFM): LLMs infer semantic identity rules from few-shot positive/negative pairs through task-specific prompting, which then guides a VFM (\eg, DINO) to extract ID-discriminative features via \textit{dynamic visual prompts}.
By aligning LLM-derived semantic concepts with the VFM's pre-trained prior, \methodname enables generalization to novel categories, eliminating the need for dataset-specific retraining. To support evaluation, we introduce ShopID10K, a dataset of 10K object instances from e-commerce platforms, featuring multi-view images and cross-domain testing. Experiments on ShopID10K and diverse ReID benchmarks demonstrate that \methodname outperforms baselines by a clear margin on unseen categories.
Code is available at \url{https://github.com/Hzzone/VICP}.

\end{abstract}

\section{Introduction}

\begin{figure}[t]
    \centering
    \includegraphics[width=1.0\linewidth]{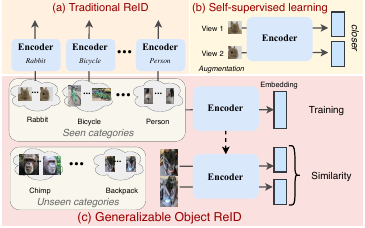}
    \vspace{-5mm}
    \caption{
        Overview of Object Re-Identification: (a) Traditional ReID methods trained the model on specific categories (\eg, rabbit, bicycle, person); (b) Self-supervised learning encourages pull together different augmentations of the same image; (c) Generalizable Object ReID extends ReID to novel categories (\eg, chimp, backpack), enabling ReID on unseen categories.
        \vspace{-3mm}
    }
    \label{fig:demo}
\end{figure}

Object Re-Identification (ReID) aims to identify and match specific instances of objects across non-overlapping camera views and scenarios, a capability critical for autonomous systems, surveillance~\cite{farsight-a-physics-driven-whole-body-biometric-system-at-large-distance-and-altitude}, and e-commerce. While extensively studied for persons~\cite{he2024instruct} and vehicles~\cite{liu2016deep,zapletal2016vehicle}, ReID applications span diverse object categories like pets~\cite{shinoda2024petface}, products~\cite{oh2016deep}, and wildlife~\cite{vcermak2024wildlifedatasets}. Unlike object classification, ReID demands distinguishing fine-grained intra-class variations—\eg, scratch on vehicles or logos for products—while remaining invariant to viewpoint, lighting, and occlusion.

Traditional approaches address challenges like visible-infrared ReID~\cite{liu2022learning,yang2022learning}, text-to-image retrieval~\cite{chen2023towards,bai2023rasa}, un/semi-supervised ReID~\cite{zhu2022pass,liu2023multiple}, domain adaptation~\cite{wei2018person}, and clothes-changing scenarios~\cite{gu2022clothes,huang2021clothing}. Yet, these methods remain category-specific: models trained on persons fail on vehicles or products, requiring costly labeled data for each new category, as shown in Fig.~\ref{fig:demo}(a). 
This specialization hinders deployment in dynamic real-world settings where novel objects (\eg, rare animal species, or emerging retail items) require rapid adaptation.

Self-supervised learning (SSL) methods like DINO~\cite{caron2021emerging,oquab2023dinov2} and MoCo~\cite{He2019MomentumCF,chen2020improved,chen2021empirical} learn representations by maximizing similarity between augmented views of an image, like contrastive loss~\cite{wu2018unsupervised} (see Fig.~\ref{fig:demo}(b)). While SSL reduces annotation needs and improves generalization to unseen domains, its objective—preserving semantic consistency—aligns poorly with ReID's core requirement: capturing fine-grained, identity-sensitive features. For instance, person ReID~\cite{he2021transreid,he2024instruct,zhu2022pass,chen2023beyond} requires distinguishing subtle differences in body shape or accessory, while pet ReID~\cite{shinoda2024petface} relies on unique fur patterns or facial markings. 
Consequently, SSL-trained models, usually beneficial for classification/detection/segmentation, often overlook these discriminative local cues, leading to suboptimal ReID performance.

A critical question emerges, as shown in Fig.~\ref{fig:demo}(c): \textit{How can we build a ReID model that generalizes to arbitrary object categories without dataset-specific training?}
Vision foundation models~(VFMs), \eg DINOv2~\cite{oquab2023dinov2} and CLIP~\cite{radford2021learning}, offer strong visual priors, but their general-purpose features lack task-specific adaptation for ReID. In contrast, large language models (LLMs)~\cite{doveh2024towards,zong2025vlicl} excel at in-context learning~\cite{dong2022survey}—extracting task rules from minimal examples.
We propose that unifying these paradigms can unlock generalization: LLMs can infer identity-discriminative rules from few-shot examples, while VFMs can localize and encode fine-grained visual traits—leading to visual in-context prompting~(\methodname), a unified ReID framework.

Specifically, in-context learning~\cite{dong2022survey} enables models to solve tasks by conditioning on example input-output prompts without parameter updates. For ReID, this means providing a model with contextual pairs (\eg, positive pairs of the same instance and negative pairs of similar but distinct objects) to infer identity-sensitive attributes. For instance, given images of handbags, an LLM could deduce that ``matching stitching patterns and logo placements'' define identity, while ``color variations under different lighting'' are irrelevant. This semantic reasoning can dynamically guide the VFM to focus on task-critical features.

On the other hand, to emphasize ReID-specific traits of pre-trained VFMs, our framework translates LLM-derived semantic rules into dynamic visual prompts—task-specific instructions generated from in-context visual example pairs.
For instance, given a few-shot positive pair (\eg, two images of the same handbag under different viewpoints) and a negative pair (\eg, two visually similar but distinct handbags), the LLM analyzes their relationships to infer identity-critical attributes (\eg, ``focus on logo placement and stitching patterns while disregarding lighting variations").
These inferred rules are then mapped into visual prompts~\cite{jia2022visual} that tune the VFM's feature extraction process.
Unlike text-based prompting, our visual prompts are derived directly from the input example pairs, enabling the model to prioritize fine-grained local features (\eg, textures, shapes) over globally invariant semantics.
This adaptation preserves the VFM's generalization while aligning it with ReID, achieving strong generalization without any parameter updates.

Despite progress in domain-specific ReID, generalizable object ReID remains underexplored. 
In this paper, we systematically establish baselines for generalizable object ReID, including self-supervised models, vision foundation models, and their adaptations.
Unlike limited categories~\cite{Market1501,MSMT17} or well-controled conditions~\cite{kotar2023these}, we further introduce ShopID10K, a dataset with instance labels curated from e-commerce platforms, comprising 10K instances across 34 daily-life categories (bag, shoes, bicycle, \etc), featuring multi-view images, occlusions, and high inter-class similarity (\eg, near-identical products differing only in logos). This benchmark enables the rigorous evaluation of cross-category generalization under real-world conditions.

Our contributions are summarized as follows:
\begin{itemize}
[leftmargin=*]
\setlength{\itemsep}{0pt} 
\setlength{\parskip}{0pt}%
\setlength{\topsep}{0pt}%
\item We define a novel task, Generalizable Object ReID, that requires ReID model to adapt to unseen categories using only a few examples.
\item We propose a unified ReID framework, \methodname, where LLMs infer identity rules from few-shot pairs, and dynamic prompts adapt VFMs for fine-grained ReID.
\item We release ShopID10K, a benchmark for evaluating cross-category ReID generalization, fostering research in scalable ReID systems.
\item Our method outperforms self-supervised and few-shot baselines by 4\% mAP on ShopID10K and standard datasets (MSMT17, VeRi-776), achieving state-of-the-art performance with minimal prompts.
\end{itemize}

\begin{figure*}[t]
    \centering
    \includegraphics[width=1.00\linewidth]{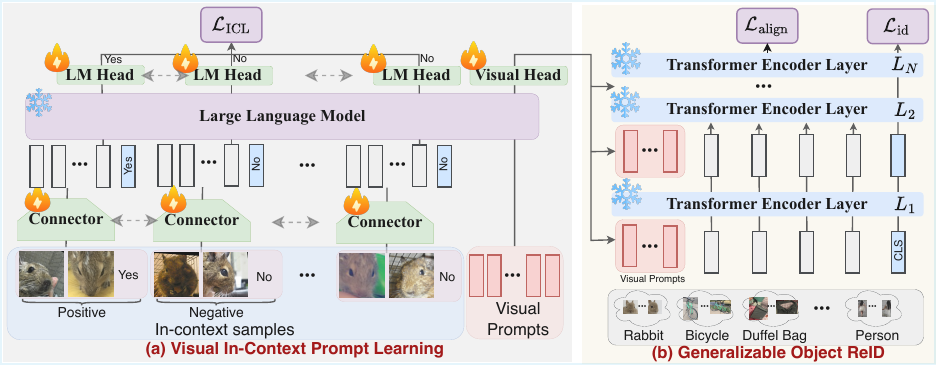}
    \vspace{-5mm}
    \caption{
    Overview of our proposed framework. a) In-Context Visual Prompt Generation: Given few-shot input pairs, a frozen LLM and a trainable connector/LM head process the context to infer identity-discriminative rules. The visual head generates learnable visual prompts conditioned on the input pairs.
    b) Generalizable Object ReID: The generated visual prompts are injected into each layer of a vision transformer (\eg, DINOv2), dynamically modulating self-attention and feature aggregation to prioritize ID-sensitive local patterns (\eg, textures, logos) while suppressing irrelevant variations (\eg, viewpoint, illumination). 
    \vspace{-3mm}
    \label{fig:framework}
    }
\end{figure*}

\section{Related Work}

\paragraph{Self-supervised Learning.}
Self-supervised learning is a powerful paradigm for learning meaningful representations without manual annotations. Contrastive learning methods, such as MoCo \cite{He2019MomentumCF} and SimCLR~\cite{chen2020simple}, construct positive pairs through data augmentation and maximize agreement between them. Subsequent works improve discriminativity of representations via clustering~\cite{huang2022learning,caron2020unsupervised}, hard negative~\cite{robinson2020contrastive} or data augmentation~\cite{wang2022contrastive}. Masked image modeling (MIM) methods~\cite{he2021masked,xie2022simmim,bao2021beit} like MAE~\cite{he2021masked} focus on reconstructing masked regions, yet their representations are less discriminative than contrastive methods for fine-grained tasks. Foundation models like DINOv2 \cite{oquab2023dinov2}, pre-trained on massive datasets, demonstrate exceptional generalization in downstream tasks~\cite{ye2024biggait,amir2021deep,jose2024dinov2,wei2025dreamrelation,xiao}. 
However, despite their semantic awareness, these representations often fail at instance-level retrieval due to the lack of explicit ID supervision.
Our method addresses this gap by leveraging LLM-guided in-context learning to inject ID-sensitive priors into VFMs.

\paragraph{Object Re-Identification.}
Traditional object ReID methods~\cite{ag-reid-2023-aerial-ground-person-re-identification-challenge-results,distilling-clip-with-dual-guidance-for-learning-discriminative-human-body-shape-representation,learning-clothing-and-pose-invariant-3d-shape-representation-for-long-term-person-re-identification,sapiensid-foundation-for-human-recognition,hamobe-hierarchical-and-adaptive-mixture-of-biometric-experts-for-video-based-person-reid,liu2016deep,zapletal2016vehicle} are category-specific, requiring dedicated training for pedestrians or vehicles. While variants like unsupervised ReID~\cite{zhu2022pass} and cross-modal ReID \cite{chen2023towards} address label scarcity or modality gaps, they remain confined to predefined categories. Deploying these methods to novel categories (\eg, pets~\cite{shinoda2024petface} or products~\cite{oh2016deep}) necessitates laborious data collection and retraining. 
This limitation highlights the need for a unified framework capable of generalizing across unseen object categories.
Indeed, our method allows a single model to adaptively identify object instances from unseen categories using only a few exemplars.

\paragraph{In-context Learning.}
In-context learning (ICL)~\cite{dong2022survey}, popularized by LLMs \cite{zhao2023survey}, where a model can rapidly adapt to new tasks during inference by conditioning on a small number of examples (\ie, prompts). Recently, researchers have begun to explore this concept in multi-modal scenarios~\cite{alayrac2022flamingo,awadalla2023openflamingo,zong2025vlicl,doveh2024towards}. In these multi-modal in-context learning frameworks, the model is provided with pairs of images and text prompts, guiding it to perform specific tasks, \eg, image classification/OCR~\cite{zong2025vlicl} and visual question answering~\cite{doveh2024towards} by leveraging cross-modal cues.
We leverage the ICL to understand image/label pairs and dynamically generate visual prompts for VFMs toward ID-discriminative features.

\paragraph{Generalizable Object ReID Datasets.}
Generalizable ReID remains underexplored due to the absence of diverse benchmarks. The sole prior work~\cite{kotar2023these} introduces a lab-controlled dataset with 180 instances from 50 categories, which  removes background to enhance self-supervised learning models. 
However, its limited scale (180 instances) and artificial environments hinder practical evaluation. 
Moreover, precise object segmentation is often infeasible in real-world scenarios. 
Our work addresses these gaps with a large-scale dataset (10K instances) and a novel visual prompting framework for generalizable object ReID.
\section{The Proposed Approach}

\subsection{Problem Formulation}

Since we propose a new task, it is necessary to first formulate our task. Traditional ReID methods require category-specific training with extensive labeled data, limiting generalization to new object categories, while self-supervised models learn generic semantics, lacking fine-grained ID patterns.

In generalizable object ReID, we aim to learn a universal feature extractor \(\phi(\cdot)\) that adapts to unseen object categories using only a small support set of examples. During training, the model is exposed to a base dataset \(\mathcal{D}_{\text{base}} = \{(\vct{x}_i, y_i, \vct{c}_i)\}\), where each image \(\vct{x}_i\) belongs to an instance-level identity \(\vct{y}_i\) within a known object category \(\vct{c}_i \in \mathcal{C}_{\text{base}}\) (\eg, backpacks, shoes). 
Critically, identities are unique within their categories, and instances across different categories are inherently distinct.

At test time, for a novel category \(c' \in \mathcal{C}_{\text{novel}}\) (disjoint from \(\mathcal{C}_{\text{base}}\)), the model is provided with a support set \(\mathcal{S} = \{(\vct{x}_i, \vct{x}_j, y_{ij})\}\) containing labeled positive (\(y_{ij} = 1\)) and negative (\(y_{ij} = 0\)) pairs, where labels indicate whether \(\vct{x}_j\) and \(\vct{x}_j\) share the same instance ID. 
The feature extractor \(\phi(\vct{x} | \mathcal{S})\) produces discriminative representations for query/gallery images from \(c'\), enabling accurate similarity computation (\eg, cosine distance) solely conditioned on \(\mathcal{S}\).
Our goal is to retrieve another image of the same instance, given a query image of one object instance.
With this goal, it's not necessary to compare instances from different categories because cross-category comparison can never retrieve the correct result. 
The key challenge of object ReID is to distinguish subtle differences across different instances belonging to the same category while being invariant to background, lighting, or pose.

Therefore, a foundational assumption is category-aware inference: during deployment, the object category \(c'\) is known a priori (\eg, via a pre-trained detector), ensuring that cross-instance comparisons are restricted to within-category pairs. This aligns with real-world ReID pipelines, where an upstream detection stage first filters candidates to a specific category (\eg, shoes), drastically reducing the search space and avoiding redundant cross-category matches (\eg, comparing a shoe to a backpack). Consequently, our framework does not need to handle cross-category ambiguity, as identities are only compared within the same category.  

Similar to few-shot learning paradigms~\cite{snell2017prototypical,wang2019panet}, generalization to novel categories is achieved via dynamic conditioning on the support set \(\mathcal{S}\), which guides \(\phi(\cdot)\) to emphasize category-specific discriminative features (\eg, shoe tread patterns, bag stitching details). This parameter-free adaptation mirrors real-world scalability requirements, where deploying ReID systems for new categories can avoid costly retraining.

\subsection{In-Context Visual Prompt Generation}
In-context learning (ICL)~\cite{dong2022survey} enables models to infer task-specific rules from provided examples without parameter updates. Unlike traditional supervised learning, ICL leverages the inherent reasoning capability of pre-trained LLMs to dynamically adapt to new tasks through sequential prompting. For ReID, this paradigm offers a critical advantage: the ability to encode \textit{identity-discriminative priors} directly from visual context (\eg, ``match objects based on logo details'') while avoiding costly fine-tuning of large vision models.  

As shown in Fig.~\ref{fig:framework}(a), our method employs a frozen LLM~(\eg, LLaMA~\cite{touvron2023llama}) to process in-context example pairs and generate semantic guidance for ReID. Given a support set \(\mathcal{S} = \{(\vct{x}_i, \vct{x}_j, y_{ij})\}\) of positive (\(y_{ij}=1\)) and negative (\(y_{ij}=0\)) image pairs, we first encode each image \(\vct{x}_i\) into visual tokens using a pre-trained vision encoder (\eg, DINOv2~\cite{oquab2023dinov2}). To mitigate computational overhead from excessive tokens (\eg, too many input pairs), we introduce a Query-based Connector (Q-Former), inspired by BLIP-2~\cite{li2023blip}, which compresses each image into a fixed set of \(N\) latent tokens. For a pair \((\vct{x}_i, \vct{x}_j)\), the connector outputs two compressed token sequences \(\mathbf{I}_i, \mathbf{I}_j \in \mathbb{R}^{N \times d}\), which are concatenated into a unified sequence:  
\begin{align}
\mathbf{T}_{ij} = [\mathbf{I}_i; \mathbf{I}_j; \mathbf{L}_{ij}] \in \mathbb{R}^{(2N+1) \times d},
\end{align}
where \(\mathbf{L}_{ij}\in \mathbb{R}^{d}\) is a learnable embedding indicating the pair's label (positive/negative). For \(K\) pairs, the full input sequence becomes \(\mathbf{T}_{\text{ctx}} = [\mathbf{T}_{ij}^{(1)}; \dots; \mathbf{T}_{ij}^{(K)}]\), forming a contextualized prompt. 
Instead of adopting LLaVA-style projection~\cite{liu2024visual}, the Q-Former can significantly reduce the number of visual tokens, $N$ for Q-former can be much smaller than the number of feature tokens of the pre-trained model in LLaVA, \eg, 256 for a single image in ViT.

The LLM processes \(\mathbf{T}_{\text{ctx}}\) to predict the next token in an auto-regressive manner. 
Crucially, we mask the loss to only supervise the label tokens \(\mathbf{L}_{ij}\), preserving the LLM's pre-trained semantic knowledge while aligning it for ReID. The training loss for a sequence of \(K\) pairs is:  
\begin{align}
\mathcal{L}_{\text{ICL}} = -\sum_{k=1}^K \log P\left(\mathbf{L}_{ij}^{(k)} \mid \mathbf{T}_{\text{ctx}}^{<k}, \mathbf{I}_i^{(k)}, \mathbf{I}_j^{(k)}\right),
\end{align}
where \(\mathbf{T}_{\text{ctx}}^{<k}\) denotes all preceding pairs in the context. This forces LLM to reason across multiple pairs, identifying discriminative patterns (\eg, ``logo consistency matters more than color'') that generalize beyond individual examples.  

To generate adaptive prompts that encode ID-discriminative knowledge, we append a set of $M$ learnable visual prompt tokens \(\mathbf{P}_{\text{learn}} \in \mathbb{R}^{M \times d}\) to the end of the input sequence \(\mathbf{T}_{\text{ctx}}\). These tokens, initialized randomly, are jointly optimized with the connector and LM head during training. The full input to the LLM becomes:  
\begin{align}
\mathbf{T}_{\text{full}} = [\mathbf{T}_{\text{ctx}}; \mathbf{P}_{\text{learn}}] \in \mathbb{R}^{(2NK + K + M) \times d}.
\end{align}

The LLM processes \(\mathbf{T}_{\text{full}}\) to contextualize the learnable prompts with the provided example pairs.
The final hidden states corresponding to \(\mathbf{P}_{\text{learn}}\) are then fed into a lightweight Visual Head—a two-layer MLP—to produce the task-specific visual prompts:  
\begin{align}
\mathbf{P}_{\text{task}} = \text{MLP}(\mathbf{P}_{\text{learn}} \cdot W_{\text{LLM}}^\top) \in \mathbb{R}^{M \times d_{\text{vision}}},
\end{align}
where $W_{\text{LLM}}$ projects the LLM's hidden dimension
$d$ to the vision model's feature dimension
$d_{\text{vision}}$.
These prompts \(\mathbf{P}_{\text{task}}\) implicitly encode \textit{how to compare} instances for ReID, distilling identity-sensitive cues (\eg, ``focus on texture consistency'') from the in-context pairs.

\subsection{Generalizable Object Re-Identification}

Vision foundation models like DINOv2~\cite{oquab2023dinov2}, pre-trained on large-scale datasets, learn rich visual-semantic representations that generalize across domains. However, while these models excel at high-level semantic tasks (\eg, classification or retrieval), their features lack the \textit{fine-grained discriminability} required for ReID.
For instance, DINOv2 may group images of ``backpacks'' by color or shape but fail to distinguish subtle ID-specific traits (\eg, logo placement or stitching patterns).
Directly fine-tuning such models on ReID data risks overfitting to specific categories and degrading their generalization ability. 

To adapt the pre-trained vision model for ReID, we inject the learned visual prompts \(\mathbf{P}_{\text{task}} \in \mathbb{R}^{M \times d_{\text{vision}}}\) into each transformer layer~\cite{jia2022visual}, as presented in Fig.~\ref{fig:framework}(b). Let \(\mathbf{Z}_l \in \mathbb{R}^{(H \times W + 1) \times d_{\text{vision}}}\) denote the input tokens at layer \(l\), where \(H \times W\) are spatial dimensions and \(+1\) corresponds to the [CLS] token. The prompts \(\mathbf{P}_{\text{task}}\) are concatenated with \(\mathbf{Z}_l\) to form an augmented token sequence:  
\[
\mathbf{Z}'_l = [\mathbf{Z}_l; \mathbf{P}_{\text{task}}] \in \mathbb{R}^{(H \times W + 1 + M) \times d_{\text{vision}}}.
\]  
The self-attention mechanism then computes interactions between all tokens, allowing the prompts to dynamically \textit{reweight spatial features}—\eg, amplifying regions with ID-sensitive details (logos, textures) while suppressing irrelevant areas (backgrounds, occlusions). Crucially, the original ViT parameters remain frozen; only the prompts \(\mathbf{P}_{\text{task}}\) (generated per support set \(\mathcal{S}\)) modulate the feature space.  

To train the framework, we propose two loss functions that preserve the vision model's generalization while enhancing ID information.

\begin{table}[t]
\scalebox{1.0}{
\resizebox{0.95\textwidth}{!}{
\begin{tabular}{r@{\extracolsep{5pt}}r@{\extracolsep{5pt}}r@{\extracolsep{5pt}}r@{\extracolsep{5pt}}r@{\extracolsep{5pt}}r}
\toprule
\multirow{2}{*}{Dataset} & \multicolumn{2}{r}{Categories} & \multirow{2}{*}{Instances} & \multirow{2}{*}{Images} & \multirow{2}{*}{Variation}\\
\cmidrule{2-3}
& Seen & Unseen & & & \\
\midrule
MVImgNet~\cite{yu2023mvimgnet} & 5 & 233 & 205K & 800K & P\\
CUTE~\cite{kotar2023these}       & 10 & 40 & 180  & 17K & L/P/B\\
PetFace~\cite{shinoda2024petface}    & 5 & 8 & 257K  & 1M & L/P/B \\
MSMT17~\cite{wei2018person}     & 1 (Person) & - & 4K & 126K & L/P/B \\
Market1501~\cite{Market1501} & 1 (Person) & - & 1.5K & 32K & L/P/B \\
VeRi-776~\cite{liu2016deep}     & 1 (Vehicle) & - & 776 & 50K & L/P/B \\
ShopeID10K~(\textbf{ours}) & 7 & 27 & 10K  & 45K & L/P/B \\
\bottomrule
\end{tabular}
}}
\vspace{-2mm}
\caption{Comparison of object ReID datasets, with variations in lighting~(L), pose~(P), and background~(B). While MVImgNet has more categories/instances, it only contains pose variations.
\vspace{-3mm}
}
\label{tab:dataset}
\end{table}

\paragraph{ReID Loss}:
We adopt triplet loss to optimize global feature discriminability.
For a mini-batch of images within the same category, we sample triplets \((\vct{x}_a, \vct{x}_p, \vct{x}_n)\), where \(\vct{x}_a\) (anchor) and \(\vct{x}_p\) (positive) share the same instance ID, and \(\vct{x}_n\) (negative) has a different ID. The loss enforces a margin \(\alpha\) between positive and negative similarities:
\begin{align}
    \mathcal{L}_{\text{ID}} = \sum_{i=1}^B 
    \max\Big(0, \alpha - \text{sim}(\phi(\vct{x}_a^i), \phi(\vct{x}_p^i)) \notag\\
    + \text{sim}(\phi(\vct{x}_a^i), \phi(\vct{x}_n^i))\Big),
\end{align}
where \(B\) is the number of triple pairs and \(\phi(\vct{x})\) is [CLS] token embedding. Triplet loss is preferred over classification losses~(\eg ArcFace~\cite{deng2019arcface}, AdaFace~\cite{kim2022adaface}) or contrastive loss~\cite{wu2018unsupervised} as it only penalizes violations of the margin constraint, imposing softer updates that preserve the pre-trained model's semantic prior, while~\cite{wu2018unsupervised,deng2019arcface,kim2022adaface} push the features to align better that may degrade the generalization ability.

\paragraph{Patch Alignment Loss}:
To further refine local feature discriminability, we compute the optimal transport (OT) distance between patch embeddings of image pairs. For a pair \((\mat{x}_i, \mat{x}_j)\), let \(\mathbf{F}_i, \mathbf{F}_j \in \mathbb{R}^{(H \times W) \times d_{\text{vision}}}\) be their patch-level features (excluding [CLS]). The OT distance reflects the matching cost across local patches.
We adopt the inexact proximal point method~\cite{xie2020fast} to compute OT distance as $D_{\text{OT}}(\cdot)$.
Based on this, we define the alignment loss:
\begin{align}
\mathcal{L}_{\text{align}}
= \sum_{(\mathbf{x}_i, \mathbf{x}_j)}
\Bigl[
  \mathbb{I}(y_{ij}=1)\cdot D_{\text{OT}}(\mathbf{F}_i, \mathbf{F}_j) \notag\\- \mathbb{I}(y_{ij}=0)\cdot D_{\text{OT}}(\mathbf{F}_i, \mathbf{F}_j)
\Bigr],
\end{align}
where \(\mathbb{I}(\cdot)\) is an indicator function, 1 if condition holds and otherwise 0, and \(D_{\text{OT}}(\mathbf{F}_i, \mathbf{F}_j)\) denotes the aforementioned OT distance. This formulation encourages positive pairs, \ie, \(\mathbb{I}(y_{ij}=1)\)) to align spatially (\eg, matching similar regions such as logos), while separating negative pairs, \ie, \(\mathbb{I}(y_{ij}=0)\), thereby enhancing local feature consistency.

\subsection{Training and Inference}

The framework is trained end-to-end with a composite loss that unifies in-context learning, ID discrimination, and local feature alignment:  
\[
\mathcal{L}_{\text{total}} = \mathcal{L}_{\text{ID}} + \lambda_{\text{ICL}} \mathcal{L}_{\text{ICL}} + \lambda_{\text{align}} \mathcal{L}_{\text{align}},
\]  
where \(\lambda_{\text{*}}\) balance the contributions of each objective.  

During the testing of a novel category \(c'\), the model processes the support set \(\mathcal{S}\) through the LLM-based prompt generator to produce category-specific visual prompts \(\mathbf{P}_{\text{task}}\). These prompts are cached and reused for all query-gallery comparisons within \(c'\), incurring only a one-time computational cost for prompt generation. Subsequent feature extraction and similarity computation follow standard ReID pipelines, with the frozen VFM modulated by \(\mathbf{P}_{\text{task}}\).  
Consequently, the cached prompts enable instant deployment to new categories without re-computation, which is ideal for dynamic environments (\eg, retail inventory updates).

\section{Experiments}
\label{sec:exp}

\begin{figure}
    \centering
\includegraphics[width=1\linewidth]{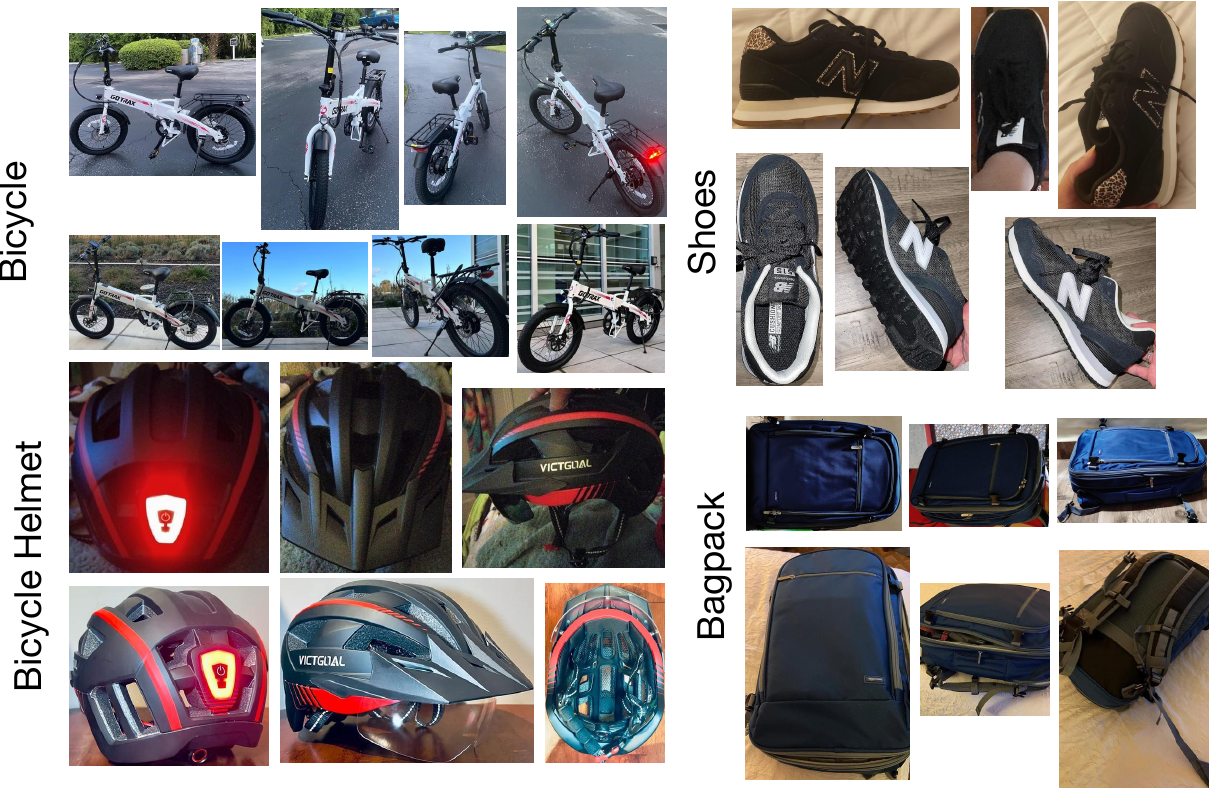}
    \caption{Examples of 4 categories, 2 instances per category, and 3-4 images per instance from our ShopID10K dataset.
    \vspace{-3mm}
    }
    \label{fig:dataset}
\end{figure}

\begin{figure*}
    \centering
    \includegraphics[width=0.85\linewidth]{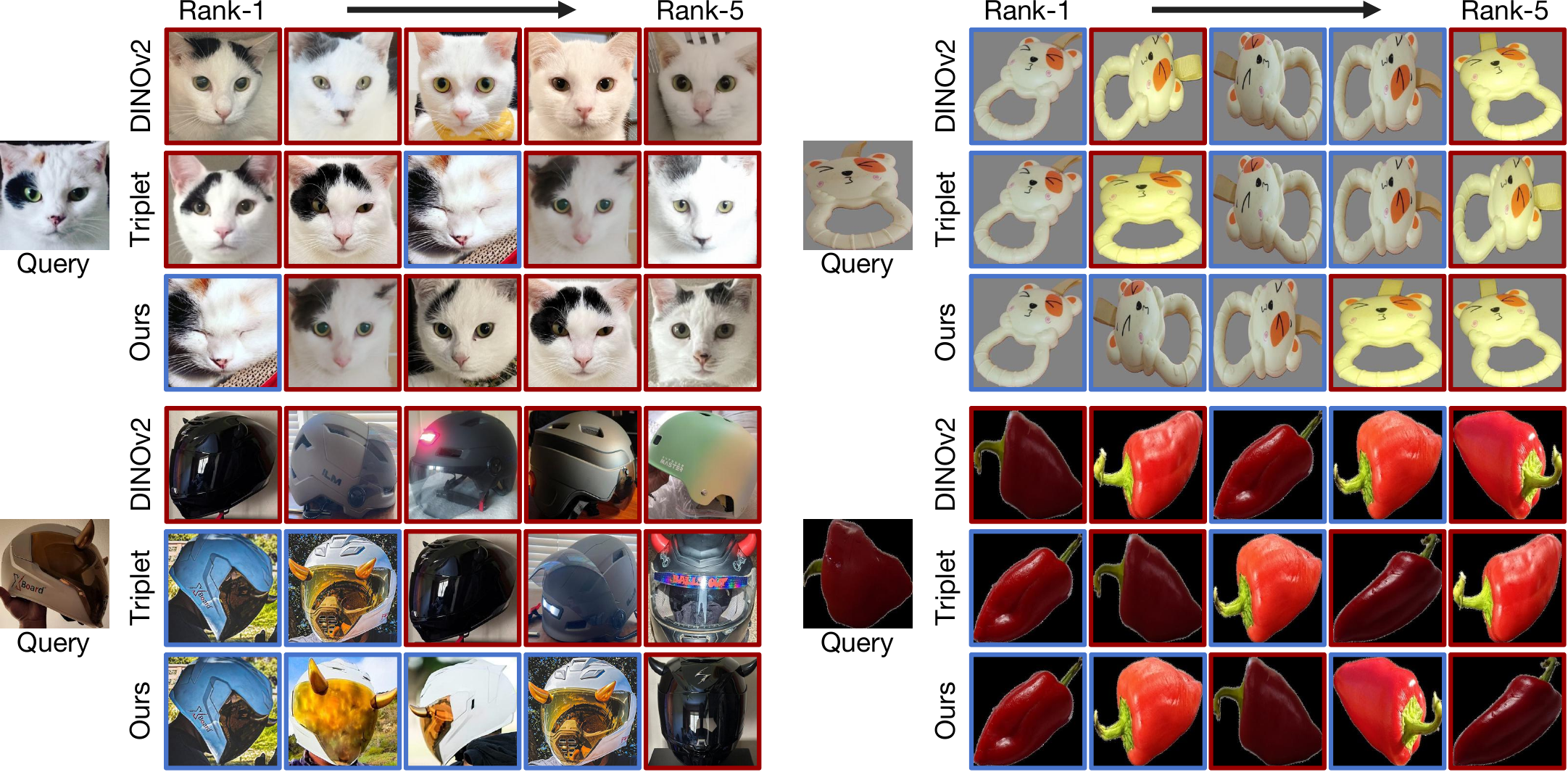}
    \caption{The ranking results of different models on PetFace/MVImageNet/ShopID10K/CUTE datasets. Blue/Red boxes indicate correct/wrong instance ID. The results show that \methodname is largely superior to baselines in finding the fine-grained information of local details.
        \vspace{-3mm}
    }
    \label{fig:vis_ret}
\end{figure*}

\paragraph{Datasets.}
We extensively evaluate on seven datasets spanning both general and specific domains, as listed in Tab.~\ref{tab:dataset}. MVImgNet~\cite{yu2023mvimgnet} provides multi-view object videos captured under controlled lighting, offering pose variations but lacking background/lighting diversity. 
To adapt it for ReID, we employ GroundingDINO~\cite{liu2023grounding} for detection and SAM 2~\cite{ravi2024sam2} to remove the background. 
We then select the four most different frames, focusing on intrinsic ID characteristics. 
CUTE~\cite{kotar2023these} captures lab-controlled images with varying illuminations/backgrounds per object, yet its limited scale (180 instances) restricts practical use. 
PetFace~\cite{shinoda2024petface} aggregates 257K pet images from Internet.  
For domain-specific baselines, MSMT17~\cite{wei2018person}, Market1501~\cite{Market1501} and VeRi-776~\cite{liu2016deep} represent person/vehicle ReID benchmarks,
although they only contain one category.

\paragraph{ShopeID10K.}
To establish a comprehensive benchmark for generalizable object ReID, we first define 34 daily object categories, including backpack, bicycle, \etc. For each category, we collect product images from Amazon customer reviews by searching category keywords. Crucially, we treat all images uploaded by the same reviewer for a specific product as sharing the same instance ID, simulating real-world scenarios where multiple images of an object instance are captured by everyday users. After detection and filtering, we ensure each instance has at least 3 images, leading to 10K instances and 45K images. The key advantage of ShopeID10K is the diversity, as shown in Fig.~\ref{fig:dataset}. 
Each instance exhibits natural variations in lighting, occlusion, pose, and background.

\paragraph{Implementation Details.}
All experiments are conducted on two H100 GPUs with a fixed learning rate of $10^{-4}$, weight decay of $10^{-4}$, $\beta_1=0.9$ and $\beta_2=0.99$.
The margin for triplet loss is set to $0.1$. We use pre-trained DINOv2 ViT-small~\cite{dosovitskiy2020image} as the backbone.
We train the model with the batch size $256$ for 10 epochs with the image size of $224\times 224$ and random horizontal flip as data augmentation. During training, we randomly sample 64 positive/negative pairs. The number of visual tokens for Q-former is 32.

\subsection{Qualitative Results} 
In Fig.~\ref{fig:vis_ret}, we visualize retrieval results of unsupervised DINOv2, Triplet+, and our methods on ShopID10K. DINOv2 predominantly retrieves images sharing shape similarity or semantic attributes (\eg, matching object categories) but fails to prioritize identity-specific features, resulting in frequent false positives. In contrast, triplet fine-tuning remarkably mitigates such errors by refining the embedding space to emphasize discriminative ID cues.  
These qualitative comparisons substantiate the superior discriminative capability of our method: it consistently retrieves identity-consistent instances across extreme variations in viewpoint and illumination while suppressing semantically similar distractors.

\subsection{Quantitative Results} 
We conduct extensive evaluations under the proposed generalizable object ReID paradigm. A subset of categories from each dataset is selected as base categories for training, and the rest are novel categories for testing. We repeat this process 5 times and report the average across different runs.

\begin{table}[]
    \centering
    \scalebox{0.8}{
    \begin{tabular}{lrrrrr}
    \toprule
    & \multicolumn{2}{c}{\textbf{Verification}} & &\multicolumn{2}{c}{\textbf{Identification}} \\
    \cmidrule{2-3}\cmidrule{5-6}
     & AUC & ACC & & mAP & Top-1 \\
     \midrule
    \demph{PetFace~\cite{shinoda2024petface}} & \demph{92.1} & \demph{-} & & \demph{-} & \demph{-} \\
    \demph{MegaDescriptor~\cite{vcermak2024wildlifedatasets}} & \demph{83.7} & \demph{-} & & \demph{-} & \demph{-} \\
    \demph{Supervised} & \demph{95.5} & \demph{89.3} & & \demph{57.7} & \demph{56.3} \\
    \midrule
    CLIP~\cite{radford2021learning} & 71.5 & 64.6 & & 7.1 & 4.4 \\
    OpenCLIP~\cite{ilharco_gabriel_2021_5143773} & 73.4 & 67.3 & & 7.6 & 5.8 \\
    DINO~\cite{caron2021emerging} & 82.2 & 74.7 & & 19.7 & 16.5 \\
    DreamSim~\cite{fu2023dreamsim} & 84.2 & 76.4 & & 17.9 & 14.9 \\
    Unicom~\cite{anxiang_2023_unicom} & 73.3 & 67.2 & & 7.0 & 5.4 \\
    I-JEPA~\cite{assran2023self} & 74.0 & 67.1 & & 10.9 & 8.7 \\
    DINOv2~\cite{jose2024dinov2} & 71.6 & 65.9 & & 6.5 & 5.1 \\
    \midrule
    Arcface~\cite{deng2019arcface} & 89.1 & 81.5 & & 46.6 & 44.3 \\
    Adaface~\cite{kim2022adaface} & 89.3 & 82.4 & & 46.9 & 44.4 \\
    SCL~\cite{wu2018unsupervised,khosla2020supervised} & 91.1 & 83.1 & & 46.3 & 43.4 \\
    Triplet~\cite{balntas2016learning} & 91.7 & 84.9 & & 48.2 & 45.9 \\
    Triplet+~\cite{balntas2016learning} & \underline{92.5} & \underline{85.6} & & \underline{49.8} & \underline{47.7} \\
    \midrule
    \methodname~(\textbf{ours}) & \textbf{93.5} & \textbf{86.0} & & \textbf{51.2} & \textbf{49.7} \\
    \bottomrule
    \end{tabular}
    }
    \vspace{-2mm}
    \caption{\small Results on PetFace~\cite{shinoda2024petface}. [Key: \textbf{First}, \underline{Second}]
     \vspace{-3mm}
        }
    \label{tab:petface}
\end{table}

\paragraph{Baselines.}
There are three types of baselines for comparison: (1) fully supervised models, which are trained on labeled data encompassing all target categories (including both seen and unseen classes during training). These models establish the empirical performance upper bound for ReID systems. (2) unsupervised or weakly supervised representation learning models~\cite{radford2021learning,ilharco_gabriel_2021_5143773,caron2021emerging,fu2023dreamsim,anxiang_2023_unicom,kotar_are_2023,assran2023self,jose2024dinov2}, such as DINO~\cite{jose2024dinov2} and CLIP~\cite{radford2021learning}, dedicated to extracting generalizable visual features. For example, for downstream tasks like image retrieval, and Unicom~\cite{anxiang_2023_unicom} proposes clustering-based feature refinement. (3) To build strong baselines, we fine‑tune DINOv2 on the base categories with learnable visual prompts~\cite{jia2022visual}, augmented with metric loss functions to encourage discriminative representations, \ie, ArcFace~\cite{deng2019arcface}, AdaFace~\cite{kim2022adaface} and supervised contrastive learning~(SCL)~\cite{khosla2020supervised}.
In particular, Triplet+ fine-tunes the model by incorporating few-shot examples based on the model trained with triplet loss, enabling targeted optimization of the ReID framework.

\paragraph{Results on PetFace~(Tab.~\ref{tab:petface}).}
PetFace~\cite{shinoda2024petface} collects multiple IDs of the same pet from web sources, comprising 13 categories, 170K unique IDs, and 1 million images.
PetFace adopts two evaluation metrics aligned with face recognition paradigms: (1) \textbf{Verification} constructs positive/negative pairs per category and computes AUC/Accuracy via 10-fold cross-validation. Our experiments strictly follow PetFace's verification protocol. (2) \textbf{Identification} measures the ability to retrieve the same instance ID gallery images from query samples. Instead of using training data in PetFace, we align with standard person ReID practices. Each image in the test set is used as a query and the remaining forms the gallery, quantified by top-1 accuracy and mAP.  
Five categories are randomly selected as base categories for training, while the remainder forms novel categories for testing. 

Supervised fine-tuning on full data significantly outperforms both original PetFace and MegaDescriptor~\cite{vcermak2024wildlifedatasets} baselines, demonstrating the efficacy of large-scale pre-trained VFMs.  
Unsupervised pre-trained models exhibit suboptimal ReID performance. Notably, fine-tuning on ReID data enhances generalization even to novel categories.  

Triplet loss surpasses alternative loss functions by selectively optimizing challenging ReID pairs while preserving pre-trained representations' inherent generalizability. In contrast, other losses will harm the embeddings because they will always minimize loss.  
Triplet+ further enhances novel category performance via integrating few-shot pairs. Our models, without further parameter updates, achieves superior generalization compared to fine-tuning-based approaches.

\paragraph{Results on MVImageNe and ShopID10K~(Tab.~\ref{tab:mvi_shop}).}
Similar to PetFace, five categories are randomly selected as base categories.  
MVImageNet, constructed from multi-view object videos, primarily captures pose variations with minimal environmental complexity, resulting in relatively lower challengingness. Interestingly, even slight fine-tuning on its constrained category set yields substantial performance gains, indicating its effectiveness as a benchmark for view-invariant representation learning.  
In contrast, our newly introduced ShopID10K exhibits extreme diversity across backgrounds, lighting conditions, occlusion patterns, and viewpoints. While PetFace focuses on constrained pet facial recognition, our method observes the same trends and achieves greater improvements on ShopID10K.

\begin{table}[]
\scalebox{0.9}{
\resizebox{\textwidth}{!}{
\begin{tabular}{lrrrrrr}
\toprule
 & \multicolumn{2}{c}{\textbf{MVImageNet}}            & & \multicolumn{3}{c}{\textbf{ShopID10K}} \\
 \cmidrule{2-3}\cmidrule{5-7}
 & mAP & Rank-1 & & mAP & Rank-1 & Rank-5 \\
 \midrule
\demph{Supervised} & \demph{79.2} & \demph{88.5} & & \demph{62.6} & \demph{71.2} & \demph{89.8} \\
 \midrule
    CLIP~\cite{radford2021learning} & 39.4 & 55.6 & & 37.1 & 48.6 & 72.1 \\
    OpenCLIP~\cite{ilharco_gabriel_2021_5143773} & 41.8 & 57.8 & & 40.2 & 51.3 & 75.4\\
    DINO~\cite{caron2021emerging} & 53.0 & 69.8 & & 41.2 & 54.4 & 75.9 \\
    DreamSim~\cite{fu2023dreamsim} & 56.1 & 71.7 & & 44.4 & 56.9 & 78.7 \\
    Unicom~\cite{anxiang_2023_unicom} & 45.5 & 61.1 & & 43.8 & 54.6 & 78.2 \\
    I-JEPA~\cite{assran2023self} & 41.2 & 58.1 & & 32.7 & 45.6 & 67.3 \\
    DINOv2~\cite{jose2024dinov2} & 47.3 & 64.1 & & 34.1 & 45.7 & 67.5 \\
    \midrule
    Arcface~\cite{deng2019arcface} & 72.2 & 58.5 & & 45.3 & 56.5 & 78.0 \\
    Adaface~\cite{kim2022adaface} & 73.0 & 59.0 & & 45.6 & 57.2 & 78.7 \\
    SCL~\cite{wu2018unsupervised,khosla2020supervised} & 68.2 & 80.4 & & 46.9 & 56.6 & 79.5 \\
    Triplet~\cite{balntas2016learning} & 72.9 & 82.1 & & 50.3 & 63.1 & 80.9 \\
    Triplet+~\cite{balntas2016learning} & \underline{73.2} & \underline{84.0} & & \underline{54.8} & \underline{67.4} & \underline{85.6} \\
    \midrule
    \methodname~(\textbf{ours}) & \textbf{74.9} & \textbf{85.4} & & \textbf{58.5} & \textbf{68.4} & \textbf{87.5} \\
 \bottomrule
\end{tabular}
}
}
\vspace{-2mm}
    \caption{\small Results on MVImageNet~\cite{yu2023mvimgnet} and our  ShopID10K.\vspace{-3mm}}
    \label{tab:mvi_shop}
\end{table}

\paragraph{Results on CUTE~(Tab.~\ref{tab:cute}).}
CUTE dataset provides laboratory-controlled multi-view imagery of objects under varying illumination and pose conditions, designed as a benchmark for intrinsic object similarity metrics.
Evaluation is conducted via pairwise comparisons across instances, with performance measured using ReID metrics: mAP and top-1 accuracy. We adopt three distinct evaluation regimes: (1) In-the-wild—images contain simultaneous variations in background, pose, and illumination; (2) Illumination and (3) Pose use controlled illumination or pose variations to assess representation robustness under isolated conditions.

Conventional fine-tuning approaches suffer from severe base-category overfitting, significantly impairing in-the-wild generalization. In contrast, our method achieves consistent performance gains across all regimes, with a 2\% improvement in mAP under in-the-wild settings. These results highlight its ability to disentangle intrinsic object features from confounding environmental factors.

\paragraph{Results on Person and Vehicle ReID.}
We further validate our approach on popular person/vehicle ReID benchmarks. For person ReID, as shown in Tab.~\ref{tab:person}, our method achieves competitive performance against state-of-the-art person ReID models, where PASS~\cite{zhu2022pass} and DINO~\cite{caron2021emerging} are pre-trained on human datasets. For vehicle ReID, \methodname outperforms TransReID~\cite{he2021transreid} on the VeRi-776 dataset~\cite{liu2016deep}, achieving a higher mAP (81.2 vs.~79.6) and slightly better R1 (97.1 vs.~97.0).
While existing methods primarily optimize for dataset-specific biases, our approach enables stronger generalization to novel categories, highlighting its adaptability beyond domain-specific constraints.

\begin{table}[]
\resizebox{\textwidth}{!}{
\begin{tabular}{rcccccc}
\toprule
 & \multicolumn{2}{c}{\textbf{In-the-wild}} & \multicolumn{2}{c}{\textbf{Illumination}} & \multicolumn{2}{c}{\textbf{Pose}} \\
 \cmidrule{2-7}
 & mAP  & Top-1  & mAP  & Top-1 & mAP  & Top-1 \\
 \midrule
 CLIP~\cite{radford2021learning} & 70.7 & 59.6 & 72.2 & 71.8 & 77.8 & 95.1 \\
    OpenCLIP~\cite{ilharco_gabriel_2021_5143773} & 75.0 & 68.4 & 73.6 & 75.9 & 79.3 & 95.7\\
    DINO~\cite{caron2021emerging} & 71.6 & 57.1 & 72.8 & 68.9 & 81.1 & 97.2\\
    DreamSim~\cite{fu2023dreamsim} & 73.0 & 59.6 & 70.0 & 65.2 & 83.4 & 97.9 \\
    Unicom~\cite{anxiang_2023_unicom} & 75.1 & 69.6 & 75.3 & 77.2 & 82.4 & 97.3 \\
    I-JEPA~\cite{assran2023self} & 61.3 & 39.6 & 65.0 & 57.6 & 74.6 & 92.1 \\
    DINOv2~\cite{jose2024dinov2} & 80.5 & 74.7 & 81.3 & 80.4 & 83.4 & 96.2 \\
    \midrule
    Arcface~\cite{deng2019arcface} & 78.2 & 74.3 & 75.1 & 75.0 & 81.5 & 96.5 \\
    Adaface~\cite{kim2022adaface} & 78.4 & 74.0 & 77.6 & 78.6 & 82.0 & 96.3 \\
    SCL~\cite{wu2018unsupervised,khosla2020supervised} & 79.4 & 75.9 & 77.7 & 77.8 & 81.6 & 96.4 \\
    Triplet~\cite{balntas2016learning} & \underline{80.9} & \underline{76.5} & \underline{84.2} & \underline{85.9} & \underline{84.1} & \underline{97.5} \\
    \midrule
    \methodname~(\textbf{ours}) & \textbf{82.5} & \textbf{77.3} & \textbf{89.8} & \textbf{89.2} & \textbf{87.6} & \textbf{98.9} \\
 \bottomrule
\end{tabular}
   \vspace{-2mm}
    \caption{Results on CUTE~\cite{kotar_are_2023}.
    \vspace{-3mm}}
    \label{tab:cute}
}
\end{table}

\begin{table}
    \centering
    \scalebox{0.8}{
    \begin{tabular}{rrrrrr}
    \toprule
    & \multicolumn{2}{c}{\textbf{MSMT17}} & &\multicolumn{2}{c}{\textbf{Market1501}} \\
    \cmidrule{2-3}\cmidrule{5-6}
     & mAP & Rank-1 & & mAP & Rank-1 \\
     \midrule
    DINO~\cite{caron2021emerging} & 66.1 & 84.6 & & \underline{91.0} & \underline{96.0} \\
    BOT~\cite{luo2019bag} & 50.2 & 74.1 & & 85.9 & 94.5 \\
    MGN~\cite{wang2018learning} & 63.7 & 85.1 & & 87.5 & 95.1 \\
    SCSN~\cite{chen2020salience} & 58.5 & 83.8 & & 88.5 & 95.7 \\
    ABDNet~\cite{chen2019abd} & 60.8 & 82.3 & & 88.3 & 95.6 \\
    AAformer~\cite{zhu2023aaformer} & 63.2 & 83.6 & & 87.7 & 95.4 \\
    TransReID~\cite{he2021transreid} & 63.6 & 82.5 & & 87.4 & 94.6 \\
    PASS~\cite{zhu2022pass} & \underline{69.1} & \underline{86.5} & & \textbf{92.2} & \textbf{96.3} \\
    \midrule
    \methodname~(\textbf{ours}) & \textbf{75.3} & \textbf{89.2} & & 90.3 & 95.5 \\
    \bottomrule
    \end{tabular}
    }
    \vspace{-2mm}
    \caption{Results on person ReID datasets.
    \vspace{-3mm}}
    \label{tab:person}
\end{table}

\subsection{Ablation Study} 
We show thee ablation studies in Tab.~\ref{tab:ablation}.

\paragraph{Ablation on Different Components.} 
To evaluate our framework, we conduct incremental ablation studies starting from the unsupervised DINOv2 baseline~{\color{linkcolor} [i]} and gradually adding our components. Even minimal fine-tuning on ReID datasets~{\color{linkcolor} [ii]}—despite category mismatches—yields noticeable gains, demonstrating the strong transferability of pre-trained visual priors to identity-sensitive tasks.
Few-shot supervision~(Triplet+~{\color{linkcolor} [iii]}) brings moderate improvement, though its scalability is limited in real-world scenarios.

While LLMs exhibit promising in-context learning (ICL~{\color{linkcolor} [v]}) capabilities, performing pairwise query-gallery inference directly is computationally impractical for large-scale ReID. Instead, we utilize LLMs as auxiliary semantic guides, encoding relational structures into ICL-based visual prompts~{\color{linkcolor} [iv]}. This enhances the VFM’s cross-domain reasoning without introducing LLM inference costs at runtime.

Finally, the patch alignment loss~{\color{linkcolor} [vi]} enforces spatial consistency across discriminative regions, improving feature localization and leading to better patch-level embedding separation, which ultimately boosts ReID accuracy.

\paragraph{Ablation on Number of Examples.} 
We investigate the impact of the number of in-context examples with $K\in \{32, 64, 128\}$ examples per category. Our method achieves peak performance at 64 examples, with 32 examples yielding marginally lower results and 128 examples causing degradation. With 32 examples, the LLM struggles to capture nuanced ID discrimination patterns, while it is too long for 128 examples to learn the complex patterns.

\paragraph{Ablation on Number of Latent Tokens.}
We vary the number of latent tokens $N \in \{16, 32, 64\}$ in the Query-based connector. The best performance is achieved at $N=32$. A smaller $N$ (16) lacks expressiveness, while a larger $N$ (64) leads to longer, noisier inputs that hurt performance and increase cost. This aligns with our findings on in-context example count, indicating both overly simple and overly complex prompts degrade results.
In supplementary, we show additional ablation in occlusion robustness, cross-domain generalization, and comparisons with few-shot learning methods.

\begin{table}[t!]
    \centering
    \scalebox{0.8}{
    \begin{tabular}{lrrrrr}
    \toprule
    & \multicolumn{2}{c}{\textbf{PetFace}} & &\multicolumn{2}{c}{\textbf{ShopID10K}} \\
    \cmidrule{2-3}\cmidrule{5-6}
     & mAP & Top-1 & & mAP & Rank-1 \\
     \midrule
    \demph{Supervised} & \demph{57.7} & \demph{56.3} & & \demph{62.6} & \demph{71.2} \\
    \midrule
    {\color{linkcolor} [i]} Unsupervised~\cite{jose2024dinov2} & 6.5 & 5.1 & & 34.1 & 45.7 \\
    \quad{\color{linkcolor} [ii]}+Triplet & 48.2 & 45.9 & & 50.3 & 63.1 \\
    \quad\quad{\color{linkcolor} [iii]} Triplet+ & 49.8 & 47.7 & & 54.8 & 67.4 \\
    \quad{\color{linkcolor} [iv]}+ICL visual prompts & 50.8 & 48.6 & & 56.3 & 67.1 \\
    \quad\quad{\color{linkcolor} [v]} ICL from LLM & 42.1 & 37.5 & & 39.6 & 48.5 \\
    \quad{\color{linkcolor} [vi]}+Patch Align & 51.2 & 49.7 & & 58.5 & 68.4 \\
    \midrule
    $K$=32 & 50.4 & 47.9 & & 55.7 & 67.5 \\
    ~~~~~~64 & 51.2 & 49.7 & & 58.5 & 68.4 \\
    ~~~~~~128 & 48.8 & 47.0 & & 56.8 & 66.9 \\
    
    \midrule
    $N$=16 & 49.9 & 48.1 & & 55.9 & 67.0 \\
    ~~~~~~32 & 51.2 & 49.7 & & 58.5 & 68.4 \\
    ~~~~~~64 & 48.6 & 47.5 & & 57.1 & 67.6 \\
    \bottomrule
    \end{tabular}
    }
    \vspace{-2mm}
    \caption{Results of ablation study.
    \vspace{-3mm}}
    \label{tab:ablation}
\end{table}
\section{Conclusion}
We address the underexplored challenge of Generalizable Object Re-Identification by integrating large language models (LLMs) with vision foundation models. Key contributions include: (1) reformulating ReID as an in-context learning task with LLM-guided feature extraction from few-shot exemplars, and (2) introducing the first large-scale benchmark with 10K real-world instances across 34 categories—surpassing prior lab-controlled datasets in diversity and complexity. Our method outperforms self-supervised baselines on novel categories while requiring less labeled data than conventional ReID. In the future, we aim to tackle the harder zero-shot ReID setting without exemplars.

\clearpage
{\small
\bibliographystyle{ieee_fullname}
\bibliography{ref}

\begin{thebibliography}{10}\itemsep=-1pt

\bibitem{alayrac2022flamingo}
Jean-Baptiste Alayrac, Jeff Donahue, Pauline Luc, Antoine Miech, Iain Barr, Yana Hasson, Karel Lenc, Arthur Mensch, Katherine Millican, Malcolm Reynolds, et~al.
\newblock Flamingo: a visual language model for few-shot learning.
\newblock In {\em Advances in neural information processing systems}, 2022.

\bibitem{amir2021deep}
Shir Amir, Yossi Gandelsman, Shai Bagon, and Tali Dekel.
\newblock Deep vit features as dense visual descriptors.
\newblock {\em arXiv preprint arXiv:2112.05814}, 2(3):4, 2021.

\bibitem{anxiang_2023_unicom}
Xiang An, Jiankang Deng, Kaicheng Yang, Jiawei Li, Ziyong Feng, Jia Guo, Jing Yang, and Tongliang Liu.
\newblock Unicom: Universal and compact representation learning for image retrieval.
\newblock In {\em ICLR}, 2023.

\bibitem{assran2023self}
Mahmoud Assran, Quentin Duval, Ishan Misra, Piotr Bojanowski, Pascal Vincent, Michael Rabbat, Yann LeCun, and Nicolas Ballas.
\newblock Self-supervised learning from images with a joint-embedding predictive architecture.
\newblock In {\em Proceedings of the IEEE/CVF Conference on Computer Vision and Pattern Recognition}, pages 15619--15629, 2023.

\bibitem{awadalla2023openflamingo}
Anas Awadalla, Irena Gao, Josh Gardner, Jack Hessel, Yusuf Hanafy, Wanrong Zhu, Kalyani Marathe, Yonatan Bitton, Samir Gadre, Shiori Sagawa, et~al.
\newblock Openflamingo: An open-source framework for training large autoregressive vision-language models.
\newblock {\em arXiv preprint arXiv:2308.01390}, 2023.

\bibitem{bai2023rasa}
Yang Bai, Min Cao, Daming Gao, Ziqiang Cao, Chen Chen, Zhenfeng Fan, Liqiang Nie, and Min Zhang.
\newblock Rasa: Relation and sensitivity aware representation learning for text-based person search.
\newblock {\em arXiv preprint arXiv:2305.13653}, 2023.

\bibitem{balntas2016learning}
Vassileios Balntas, Edgar Riba, Daniel Ponsa, and Krystian Mikolajczyk.
\newblock Learning local feature descriptors with triplets and shallow convolutional neural networks.
\newblock In {\em Bmvc}, volume~1, page~3, 2016.

\bibitem{bao2021beit}
Hangbo Bao, Li Dong, Songhao Piao, and Furu Wei.
\newblock Beit: Bert pre-training of image transformers.
\newblock {\em arXiv preprint arXiv:2106.08254}, 2021.

\bibitem{caron2020unsupervised}
Mathilde Caron, Ishan Misra, Julien Mairal, Priya Goyal, Piotr Bojanowski, and Armand Joulin.
\newblock Unsupervised learning of visual features by contrasting cluster assignments.
\newblock 2020.

\bibitem{caron2021emerging}
Mathilde Caron, Hugo Touvron, Ishan Misra, Herv{\'e} J{\'e}gou, Julien Mairal, Piotr Bojanowski, and Armand Joulin.
\newblock Emerging properties in self-supervised vision transformers.
\newblock In {\em Proceedings of the IEEE/CVF international conference on computer vision}, pages 9650--9660, 2021.

\bibitem{vcermak2024wildlifedatasets}
Vojt{\v{e}}ch {\v{C}}erm{\'a}k, Lukas Picek, Luk{\'a}{\v{s}} Adam, and Kostas Papafitsoros.
\newblock Wildlifedatasets: An open-source toolkit for animal re-identification.
\newblock In {\em Proceedings of the IEEE/CVF Winter Conference on Applications of Computer Vision}, pages 5953--5963, 2024.

\bibitem{chen2023towards}
Cuiqun Chen, Mang Ye, and Ding Jiang.
\newblock Towards modality-agnostic person re-identification with descriptive query.
\newblock In {\em Proceedings of the IEEE/CVF conference on computer vision and pattern recognition}, pages 15128--15137, 2023.

\bibitem{chen2019abd}
Tianlong Chen, Shaojin Ding, Jingyi Xie, Ye Yuan, Wuyang Chen, Yang Yang, Zhou Ren, and Zhangyang Wang.
\newblock Abd-net: Attentive but diverse person re-identification.
\newblock In {\em Proceedings of the IEEE/CVF international conference on computer vision}, pages 8351--8361, 2019.

\bibitem{chen2020simple}
Ting Chen, Simon Kornblith, Mohammad Norouzi, and Geoffrey Hinton.
\newblock A simple framework for contrastive learning of visual representations.
\newblock In {\em International conference on machine learning}, pages 1597--1607. PmLR, 2020.

\bibitem{chen2023beyond}
Weihua Chen, Xianzhe Xu, Jian Jia, Hao Luo, Yaohua Wang, Fan Wang, Rong Jin, and Xiuyu Sun.
\newblock Beyond appearance: a semantic controllable self-supervised learning framework for human-centric visual tasks.
\newblock In {\em Proceedings of the IEEE/CVF conference on computer vision and pattern recognition}, pages 15050--15061, 2023.

\bibitem{chen2020improved}
Xinlei Chen, Haoqi Fan, Ross Girshick, and Kaiming He.
\newblock Improved baselines with momentum contrastive learning.
\newblock {\em arXiv preprint arXiv:2003.04297}, 2020.

\bibitem{chen2020salience}
Xuesong Chen, Canmiao Fu, Yong Zhao, Feng Zheng, Jingkuan Song, Rongrong Ji, and Yi Yang.
\newblock Salience-guided cascaded suppression network for person re-identification.
\newblock In {\em Proceedings of the IEEE/CVF conference on computer vision and pattern recognition}, pages 3300--3310, 2020.

\bibitem{chen2021empirical}
Xinlei Chen, Saining Xie, and Kaiming He.
\newblock An empirical study of training self-supervised vision transformers.
\newblock In {\em Proceedings of the IEEE/CVF international conference on computer vision}, pages 9640--9649, 2021.

\bibitem{deng2019arcface}
Jiankang Deng, Jia Guo, Niannan Xue, and Stefanos Zafeiriou.
\newblock Arcface: Additive angular margin loss for deep face recognition.
\newblock In {\em Proceedings of the IEEE/CVF conference on computer vision and pattern recognition}, pages 4690--4699, 2019.

\bibitem{dong2022survey}
Qingxiu Dong, Lei Li, Damai Dai, Ce Zheng, Jingyuan Ma, Rui Li, Heming Xia, Jingjing Xu, Zhiyong Wu, Tianyu Liu, et~al.
\newblock A survey on in-context learning.
\newblock {\em arXiv preprint arXiv:2301.00234}, 2022.

\bibitem{dosovitskiy2020image}
Alexey Dosovitskiy, Lucas Beyer, Alexander Kolesnikov, Dirk Weissenborn, Xiaohua Zhai, Thomas Unterthiner, Mostafa Dehghani, Matthias Minderer, Georg Heigold, Sylvain Gelly, et~al.
\newblock An image is worth 16x16 words: Transformers for image recognition at scale.
\newblock {\em arXiv preprint arXiv:2010.11929}, 2020.

\bibitem{doveh2024towards}
Sivan Doveh, Shaked Perek, M~Jehanzeb Mirza, Wei Lin, Amit Alfassy, Assaf Arbelle, Shimon Ullman, and Leonid Karlinsky.
\newblock Towards multimodal in-context learning for vision \& language models.
\newblock {\em arXiv preprint arXiv:2403.12736}, 2024.

\bibitem{finn2017model}
Chelsea Finn, Pieter Abbeel, and Sergey Levine.
\newblock Model-agnostic meta-learning for fast adaptation of deep networks.
\newblock In {\em International conference on machine learning}, pages 1126--1135. PMLR, 2017.

\bibitem{fu2023dreamsim}
Stephanie Fu, Netanel Tamir, Shobhita Sundaram, Lucy Chai, Richard Zhang, Tali Dekel, and Phillip Isola.
\newblock Dreamsim: Learning new dimensions of human visual similarity using synthetic data.
\newblock {\em arXiv preprint arXiv:2306.09344}, 2023.

\bibitem{gu2022clothes}
Xinqian Gu, Hong Chang, Bingpeng Ma, Shutao Bai, Shiguang Shan, and Xilin Chen.
\newblock Clothes-changing person re-identification with rgb modality only.
\newblock In {\em Proceedings of the IEEE/CVF conference on computer vision and pattern recognition}, pages 1060--1069, 2022.

\bibitem{xiao}
Xiao Guo, Xiaohong Liu, Iacopo Masi, and Xiaoming Liu.
\newblock Language-guided hierarchical fine-grained image forgery detection and localization.
\newblock In {\em International Journal of Computer Vision}, December 2024.

\bibitem{he2021masked}
Kaiming He, Xinlei Chen, Saining Xie, Yanghao Li, Piotr Doll{\'a}r, and Ross Girshick.
\newblock Masked autoencoders are scalable vision learners.
\newblock {\em CVPR}, 2022.

\bibitem{He2019MomentumCF}
Kaiming He, Haoqi Fan, Yuxin Wu, Saining Xie, and Ross~B. Girshick.
\newblock Momentum contrast for unsupervised visual representation learning.
\newblock {\em 2020 IEEE/CVF Conference on Computer Vision and Pattern Recognition (CVPR)}, pages 9726--9735, 2019.

\bibitem{he2021transreid}
Shuting He, Hao Luo, Pichao Wang, Fan Wang, Hao Li, and Wei Jiang.
\newblock Transreid: Transformer-based object re-identification.
\newblock In {\em Proceedings of the IEEE/CVF international conference on computer vision}, pages 15013--15022, 2021.

\bibitem{he2024instruct}
Weizhen He, Yiheng Deng, Shixiang Tang, Qihao Chen, Qingsong Xie, Yizhou Wang, Lei Bai, Feng Zhu, Rui Zhao, Wanli Ouyang, et~al.
\newblock Instruct-reid: A multi-purpose person re-identification task with instructions.
\newblock In {\em Proceedings of the IEEE/CVF Conference on Computer Vision and Pattern Recognition}, pages 17521--17531, 2024.

\bibitem{huang2021clothing}
Yan Huang, Qiang Wu, JingSong Xu, Yi Zhong, and ZhaoXiang Zhang.
\newblock Clothing status awareness for long-term person re-identification.
\newblock In {\em Proceedings of the IEEE/CVF International Conference on Computer Vision}, pages 11895--11904, 2021.

\bibitem{huang2022learning}
Zhizhong Huang, Jie Chen, Junping Zhang, and Hongming Shan.
\newblock Learning representation for clustering via prototype scattering and positive sampling.
\newblock {\em IEEE Transactions on Pattern Analysis and Machine Intelligence}, 45(6):7509--7524, 2022.

\bibitem{ilharco_gabriel_2021_5143773}
Gabriel Ilharco, Mitchell Wortsman, Ross Wightman, Cade Gordon, Nicholas Carlini, Rohan Taori, Achal Dave, Vaishaal Shankar, Hongseok Namkoong, John Miller, Hannaneh Hajishirzi, Ali Farhadi, and Ludwig Schmidt.
\newblock Openclip, July 2021.
\newblock If you use this software, please cite it as below.

\bibitem{jia2022visual}
Menglin Jia, Luming Tang, Bor-Chun Chen, Claire Cardie, Serge Belongie, Bharath Hariharan, and Ser-Nam Lim.
\newblock Visual prompt tuning.
\newblock In {\em European conference on computer vision}, pages 709--727. Springer, 2022.

\bibitem{jose2024dinov2}
Cijo Jose, Th{\'e}o Moutakanni, Dahyun Kang, Federico Baldassarre, Timoth{\'e}e Darcet, Hu Xu, Daniel Li, Marc Szafraniec, Micha{\"e}l Ramamonjisoa, Maxime Oquab, et~al.
\newblock Dinov2 meets text: A unified framework for image-and pixel-level vision-language alignment.
\newblock {\em arXiv preprint arXiv:2412.16334}, 2024.

\bibitem{khosla2020supervised}
Prannay Khosla, Piotr Teterwak, Chen Wang, Aaron Sarna, Yonglong Tian, Phillip Isola, Aaron Maschinot, Ce Liu, and Dilip Krishnan.
\newblock Supervised contrastive learning.
\newblock {\em Advances in neural information processing systems}, 33:18661--18673, 2020.

\bibitem{kim2022adaface}
Minchul Kim, Anil~K Jain, and Xiaoming Liu.
\newblock Adaface: Quality adaptive margin for face recognition.
\newblock In {\em Proceedings of the IEEE/CVF conference on computer vision and pattern recognition}, pages 18750--18759, 2022.

\bibitem{sapiensid-foundation-for-human-recognition}
Minchul Kim, Dingqiang Ye, Yiyang Su, Feng Liu, and Xiaoming Liu.
\newblock Sapiensid: Foundation for human recognition.
\newblock In {\em In Proceeding of IEEE Computer Vision and Pattern Recognition}, Nashville, TN, June 2025.

\bibitem{kotar2023these}
Klemen Kotar, Stephen Tian, Hong-Xing Yu, Dan Yamins, and Jiajun Wu.
\newblock Are these the same apple? comparing images based on object intrinsics.
\newblock In {\em Advances in Neural Information Processing Systems}, 2023.

\bibitem{kotar_are_2023}
Klemen Kotar, Stephen Tian, Hong-Xing Yu, Daniel L.~K. Yamins, and Jiajun Wu.
\newblock Are {These} the {Same} {Apple}? {Comparing} {Images} {Based} on {Object} {Intrinsics}, Nov. 2023.

\bibitem{li2023blip}
Junnan Li, Dongxu Li, Silvio Savarese, and Steven Hoi.
\newblock Blip-2: Bootstrapping language-image pre-training with frozen image encoders and large language models.
\newblock In {\em International conference on machine learning}, pages 19730--19742. PMLR, 2023.

\bibitem{farsight-a-physics-driven-whole-body-biometric-system-at-large-distance-and-altitude}
Feng Liu, Ryan Ashbaugh, Nicholas Chimitt, Najmul Hassan, Ali Hassani, Ajay Jaiswal, Minchul Kim, Zhiyuan Mao, Christopher Perry, Zhiyuan Ren, Yiyang Su, Pegah Varghaei, Kai Wang, Xingguang Zhang, Stanley Chan, Arun Ross, Humphrey Shi, Zhangyang Wang, Anil Jain, and Xiaoming Liu.
\newblock Farsight: A physics-driven whole-body biometric system at large distance and altitude.
\newblock In {\em In Proceeding of Winter Conference on Applications of Computer Vision}, Waikoloa, Hawaii, January 2024.

\bibitem{learning-clothing-and-pose-invariant-3d-shape-representation-for-long-term-person-re-identification}
Feng Liu, Minchul Kim, ZiAng Gu, Anil Jain, and Xiaoming Liu.
\newblock Learning clothing and pose invariant 3d shape representation for long-term person re-identification.
\newblock In {\em In Proceeding of International Conference on Computer Vision}, Paris, France, October 2023.

\bibitem{distilling-clip-with-dual-guidance-for-learning-discriminative-human-body-shape-representation}
Feng Liu, Minchul Kim, Zhiyuan Ren, and Xiaoming Liu.
\newblock Distilling clip with dual guidance for learning discriminative human body shape representation.
\newblock In {\em In Proceeding of IEEE Computer Vision and Pattern Recognition}, Seattle, WA, June 2024.

\bibitem{liu2024visual}
Haotian Liu, Chunyuan Li, Qingyang Wu, and Yong~Jae Lee.
\newblock Visual instruction tuning.
\newblock {\em Advances in neural information processing systems}, 36, 2024.

\bibitem{liu2022learning}
Jialun Liu, Yifan Sun, Feng Zhu, Hongbin Pei, Yi Yang, and Wenhui Li.
\newblock Learning memory-augmented unidirectional metrics for cross-modality person re-identification.
\newblock In {\em Proceedings of the IEEE/CVF conference on computer vision and pattern recognition}, pages 19366--19375, 2022.

\bibitem{liu2023multiple}
Kangning Liu, Weicheng Zhu, Yiqiu Shen, Sheng Liu, Narges Razavian, Krzysztof~J Geras, and Carlos Fernandez-Granda.
\newblock Multiple instance learning via iterative self-paced supervised contrastive learning.
\newblock In {\em Proceedings of the IEEE/CVF Conference on Computer Vision and Pattern Recognition}, pages 3355--3365, 2023.

\bibitem{liu2023grounding}
Shilong Liu, Zhaoyang Zeng, Tianhe Ren, Feng Li, Hao Zhang, Jie Yang, Chunyuan Li, Jianwei Yang, Hang Su, Jun Zhu, et~al.
\newblock Grounding dino: Marrying dino with grounded pre-training for open-set object detection.
\newblock {\em arXiv preprint arXiv:2303.05499}, 2023.

\bibitem{liu2016deep}
Xinchen Liu, Wu Liu, Tao Mei, and Huadong Ma.
\newblock A deep learning-based approach to progressive vehicle re-identification for urban surveillance.
\newblock In {\em Computer Vision--ECCV 2016: 14th European Conference, Amsterdam, The Netherlands, October 11-14, 2016, Proceedings, Part II 14}, pages 869--884. Springer, 2016.

\bibitem{luo2019bag}
Hao Luo, Youzhi Gu, Xingyu Liao, Shenqi Lai, and Wei Jiang.
\newblock Bag of tricks and a strong baseline for deep person re-identification.
\newblock In {\em Proceedings of the IEEE/CVF conference on computer vision and pattern recognition workshops}, pages 0--0, 2019.

\bibitem{ag-reid-2023-aerial-ground-person-re-identification-challenge-results}
Kien Nguyen, Clinton Fookes, Sridha Sridharan, Feng Liu, Xiaoming Liu, Arun Ross, Dana Michalski, Huy Nguyen, Debayan Deb, Mahak Kothari, Manisha Saini, Dawei Du, Scott McCloskey, Gabriel Bertocco, Fernanda Andal´o, Terrance~E. Boult, Anderson Rocha, Haidong Zhu, Zhaoheng Zheng, Ram Nevatia, Zaigham Randhawa, Sinan Sabri, and Gianfranco Doretto.
\newblock Ag-reid 2023: Aerial-ground person re-identification challenge results.
\newblock In {\em In Proceeding of International Joint Conference on Biometrics}, Ljubljana, Slovenia, September 2023.

\bibitem{oh2016deep}
Hyun Oh~Song, Yu Xiang, Stefanie Jegelka, and Silvio Savarese.
\newblock Deep metric learning via lifted structured feature embedding.
\newblock In {\em Proceedings of the IEEE conference on computer vision and pattern recognition}, pages 4004--4012, 2016.

\bibitem{oquab2023dinov2}
Maxime Oquab, Timoth{\'e}e Darcet, Th{\'e}o Moutakanni, Huy Vo, Marc Szafraniec, Vasil Khalidov, Pierre Fernandez, Daniel Haziza, Francisco Massa, Alaaeldin El-Nouby, et~al.
\newblock Dinov2: Learning robust visual features without supervision.
\newblock {\em arXiv preprint arXiv:2304.07193}, 2023.

\bibitem{radford2021learning}
Alec Radford, Jong~Wook Kim, Chris Hallacy, Aditya Ramesh, Gabriel Goh, Sandhini Agarwal, Girish Sastry, Amanda Askell, Pamela Mishkin, Jack Clark, et~al.
\newblock Learning transferable visual models from natural language supervision.
\newblock In {\em International conference on machine learning}, pages 8748--8763, 2021.

\bibitem{ravi2024sam2}
Nikhila Ravi, Valentin Gabeur, Yuan-Ting Hu, Ronghang Hu, Chaitanya Ryali, Tengyu Ma, Haitham Khedr, Roman R{\"a}dle, Chloe Rolland, Laura Gustafson, Eric Mintun, Junting Pan, Kalyan~Vasudev Alwala, Nicolas Carion, Chao-Yuan Wu, Ross Girshick, Piotr Doll{\'a}r, and Christoph Feichtenhofer.
\newblock Sam 2: Segment anything in images and videos.
\newblock {\em arXiv preprint arXiv:2408.00714}, 2024.

\bibitem{robinson2020contrastive}
Joshua Robinson, Ching-Yao Chuang, Suvrit Sra, and Stefanie Jegelka.
\newblock Contrastive learning with hard negative samples.
\newblock {\em arXiv preprint arXiv:2010.04592}, 2020.

\bibitem{shinoda2024petface}
Risa Shinoda and Kaede Shiohara.
\newblock Petface: A large-scale dataset and benchmark for animal identification.
\newblock In {\em European Conference on Computer Vision}, pages 19--36. Springer, 2024.

\bibitem{snell2017prototypical}
Jake Snell, Kevin Swersky, and Richard Zemel.
\newblock Prototypical networks for few-shot learning.
\newblock In {\em Advances in neural information processing systems}, 2017.

\bibitem{hamobe-hierarchical-and-adaptive-mixture-of-biometric-experts-for-video-based-person-reid}
Yiyang Su, Yunping Shi, Feng Liu, and Xiaoming Liu.
\newblock Hamobe: Hierarchical and adaptive mixture of biometric experts for video-based person reid.
\newblock In {\em In Proceeding of International Conference on Computer Vision}, Honolulu, Hawaii, October 2025.

\bibitem{touvron2023llama}
Hugo Touvron, Thibaut Lavril, Gautier Izacard, Xavier Martinet, Marie-Anne Lachaux, Timoth{\'e}e Lacroix, Baptiste Rozi{\`e}re, Naman Goyal, Eric Hambro, Faisal Azhar, Aurelien Rodriguez, Armand Joulin, Edouard Grave, and Guillaume Lample.
\newblock Llama: Open and efficient foundation language models.
\newblock {\em arXiv preprint arXiv:2302.13971}, 2023.

\bibitem{wang2018learning}
Guanshuo Wang, Yufeng Yuan, Xiong Chen, Jiwei Li, and Xi Zhou.
\newblock Learning discriminative features with multiple granularities for person re-identification.
\newblock In {\em Proceedings of the 26th ACM international conference on Multimedia}, pages 274--282, 2018.

\bibitem{wang2019panet}
Kaixin Wang, Jun~Hao Liew, Yingtian Zou, Daquan Zhou, and Jiashi Feng.
\newblock Panet: Few-shot image semantic segmentation with prototype alignment.
\newblock In {\em proceedings of the IEEE/CVF international conference on computer vision}, pages 9197--9206, 2019.

\bibitem{wang2022contrastive}
Xiao Wang and Guo-Jun Qi.
\newblock Contrastive learning with stronger augmentations.
\newblock {\em IEEE transactions on pattern analysis and machine intelligence}, 45(5):5549--5560, 2022.

\bibitem{wei2018person}
Longhui Wei, Shiliang Zhang, Wen Gao, and Qi Tian.
\newblock Person transfer gan to bridge domain gap for person re-identification.
\newblock In {\em Proceedings of the IEEE conference on computer vision and pattern recognition}, pages 79--88, 2018.

\bibitem{MSMT17}
Longhui Wei, Shiliang Zhang, Wen Gao, and Qi Tian.
\newblock Person transfer gan to bridge domain gap for person re-identification.
\newblock In {\em Proceedings of the IEEE conference on computer vision and pattern recognition}, pages 79--88, 2018.

\bibitem{wei2025dreamrelation}
Yujie Wei, Shiwei Zhang, Hangjie Yuan, Biao Gong, Longxiang Tang, Xiang Wang, Haonan Qiu, Hengjia Li, Shuai Tan, Yingya Zhang, et~al.
\newblock Dreamrelation: Relation-centric video customization.
\newblock {\em arXiv preprint arXiv:2503.07602}, 2025.

\bibitem{wu2018unsupervised}
Zhirong Wu, Yuanjun Xiong, Stella~X Yu, and Dahua Lin.
\newblock Unsupervised feature learning via non-parametric instance discrimination.
\newblock In {\em Proceedings of the IEEE conference on computer vision and pattern recognition}, pages 3733--3742, 2018.

\bibitem{xie2020fast}
Yujia Xie, Xiangfeng Wang, Ruijia Wang, and Hongyuan Zha.
\newblock A fast proximal point method for computing exact wasserstein distance.
\newblock In {\em Uncertainty in artificial intelligence}, pages 433--453. PMLR, 2020.

\bibitem{xie2022simmim}
Zhenda Xie, Zheng Zhang, Yue Cao, Yutong Lin, Jianmin Bao, Zhuliang Yao, Qi Dai, and Han Hu.
\newblock Simmim: A simple framework for masked image modeling.
\newblock In {\em Proceedings of the IEEE/CVF conference on computer vision and pattern recognition}, pages 9653--9663, 2022.

\bibitem{yang2022learning}
Mouxing Yang, Zhenyu Huang, Peng Hu, Taihao Li, Jiancheng Lv, and Xi Peng.
\newblock Learning with twin noisy labels for visible-infrared person re-identification.
\newblock In {\em Proceedings of the IEEE/CVF conference on computer vision and pattern recognition}, pages 14308--14317, 2022.

\bibitem{ye2024biggait}
Dingqiang Ye, Chao Fan, Jingzhe Ma, Xiaoming Liu, and Shiqi Yu.
\newblock Biggait: Learning gait representation you want by large vision models.
\newblock In {\em Proceedings of the IEEE/CVF Conference on Computer Vision and Pattern Recognition}, pages 200--210, 2024.

\bibitem{yu2023mvimgnet}
Xianggang Yu, Mutian Xu, Yidan Zhang, Haolin Liu, Chongjie Ye, Yushuang Wu, Zizheng Yan, Chenming Zhu, Zhangyang Xiong, Tianyou Liang, et~al.
\newblock Mvimgnet: A large-scale dataset of multi-view images.
\newblock In {\em Proceedings of the IEEE/CVF conference on computer vision and pattern recognition}, pages 9150--9161, 2023.

\bibitem{zapletal2016vehicle}
Dominik Zapletal and Adam Herout.
\newblock Vehicle re-identification for automatic video traffic surveillance.
\newblock In {\em Proceedings of the IEEE conference on computer vision and pattern recognition workshops}, pages 25--31, 2016.

\bibitem{zhao2023survey}
Wayne~Xin Zhao, Kun Zhou, Junyi Li, Tianyi Tang, Xiaolei Wang, Yupeng Hou, Yingqian Min, Beichen Zhang, Junjie Zhang, Zican Dong, et~al.
\newblock A survey of large language models.
\newblock {\em arXiv preprint arXiv:2303.18223}, 2023.

\bibitem{Market1501}
Liang Zheng, Liyue Shen, Lu Tian, Shengjin Wang, Jingdong Wang, and Qi Tian.
\newblock Scalable person re-identification: A benchmark.
\newblock In {\em Proceedings of the IEEE international conference on computer vision}, pages 1116--1124, 2015.

\bibitem{zhu2022pass}
Kuan Zhu, Haiyun Guo, Tianyi Yan, Yousong Zhu, Jinqiao Wang, and Ming Tang.
\newblock Pass: Part-aware self-supervised pre-training for person re-identification.
\newblock In {\em European conference on computer vision}, pages 198--214. Springer, 2022.

\bibitem{zhu2023aaformer}
Kuan Zhu, Haiyun Guo, Shiliang Zhang, Yaowei Wang, Jing Liu, Jinqiao Wang, and Ming Tang.
\newblock Aaformer: Auto-aligned transformer for person re-identification.
\newblock {\em IEEE Transactions on Neural Networks and Learning Systems}, 2023.

\bibitem{zong2025vlicl}
Yongshuo Zong, Ondrej Bohdal, and Timothy Hospedales.
\newblock {VL}-{ICL} bench: The devil in the details of multimodal in-context learning.
\newblock In {\em The Thirteenth International Conference on Learning Representations}, 2025.

\end{thebibliography}
}

\clearpage
\clearpage
\setcounter{page}{1}
\maketitlesupplementary

\section{Additional Experiments}
\label{sec:supp_exp}

\paragraph{Robustness to occlusions}:
Robustness to pose/lighting has been validated through MVImageNet and CUTE datasets, which offer rich pose variations via multi-view videos or lab-controlled pose/lighting variations.
While ShopID10K is visually observed with occlusion variations, there are no explicit occlusion labels, making it difficult to quantitatively evaluate occlusion robustness. To address this, we leverage SAM segmentation model to generate object segmentation maps for ShopID10K and perform connected components analysis to construct a subset of occluded objects based on the disjoint regions. We then use this subset as the query to evaluate VICP and baseline methods. As shown in Tab.~\ref{tab:ablation_study_occlusion}, our method consistently outperforms baselines under occluded conditions with strong robustness to occlusions.

\begin{table}[h]
    \centering
    \begin{tabular}{c|cccc}
     & DINOv2 & Triplet & Triplet+ & VICP \\\hline
    mAP & 25.7 & 40.5 & 46.8 & \textbf{50.2} \\
    Rank‑1 & 36.1 & 52.4 & 59.2 & \textbf{61.4}
    \end{tabular}
    \caption{Results on occluded ShopID10K subset.}
    \label{tab:ablation_study_occlusion}
\end{table}

\paragraph{Cross-Domain evaluation}:
We performed cross-domain evaluation by using the models trained on PetFace/MVImageNet/CUTE to evaluate directly on ShopID10K in Tab.~\ref{tab:crossdomain}. Despite the inherent difficulty of this setting, our method consistently outperforms Triplet+, demonstrating its ability to generalize across significantly different domains. 
The smaller gains of VICP on PetFace compared to other two datasets suggest that large domain gaps constrain its generalization ability.
\begin{table}[h]
    \centering
    \vspace{-10pt}
    \begin{tabular}{c|c@{\hspace{3pt}}c@{\hspace{3pt}}c}
    & PetFace & MVImageNet & CUTE \\\hline
    Triplet+ & 40.2 & 51.6 & 47.9 \\
    VICP & \textbf{41.6} & \textbf{54.2} & \textbf{52.1}
    \end{tabular}
    \caption{Cross-domain mAP.}
    \label{tab:crossdomain}
\end{table}

\paragraph{Comparisons with few-shot learning techniques}:
Few-shot methods like prototypical or matching networks target coarse-grained tasks, \eg, classification. In contrast, ReID requires fine-grained, identity-level discrimination, limiting their utility.
We therefore evaluate the most relevant alternative—model-agnostic meta-learning (MAML)~\cite{finn2017model}. We train MAML on base categories and fine-tune it on the few-shot examples from unseen categories (same setup as Triplet+). As Tab.~\ref{tab:ablation_study_maml} shows, MAML marginally outperforms Triplet+, yet still falls short of VICP, which requires no additional fine-tuning.

\begin{table}[h]
    \centering
    \begin{tabular}{c|cccc}
    & Triplet & Triplet+ & MAML & VICP \\\hline
    mAP & 50.3 & 54.8 & 55.9 & \textbf{58.5} \\
    Rank‑1 & 63.1 & 67.4 & 68.1 & \textbf{68.4}
    \end{tabular}
    \caption{Results of MAML on ShopID10K.}
    \label{tab:ablation_study_maml}
\end{table}

\end{document}